# Image Processing in Optical Guidance for Autonomous Landing of Lunar Probe

Ding Meng [1]   Cao Yun-feng [2]   Wu Qing-xian [1]  and  Zhang Zhen [1]

[1] School of Automatic Engineering
NanJing University of Aeronautics and Astronautics
E-mail:nuaa_dm@hotmail.com

[2] Academy of Frontier Science
NanJing University of Aeronautics and Astronautics
E-mail:cyfac@nuaa.edu.cn

**Abstract**

Because of the communication delay between earth and moon, the GNC technology of lunar probe is becoming more important than ever. Current navigation technology is not able to provide precise motion estimation for probe landing control system Computer vision offers a new approach to solve this problem. In this paper, author introduces an image process algorithm of computer vision navigation for autonomous landing of lunar probe. The purpose of the algorithm is to detect and track feature points which are factors of navigation. Firstly, fixation areas are detected as sub-images and matched. Secondly, feature points are extracted from sub-images and tracked. Computer simulation demonstrates the result of algorithm takes less computation and fulfils requests of navigation algorithm.

## I. INTRODUCTION

In paper[1], the status of China's deep space exploration plan is introduced including CE-1 lunar orbiter, the china's subsequent Lunar Exploration Program. It is an important purpose in the second stage of china lunar exploration to land accurately of probe on the moon's surface.

The guidance-navigation-control(GNC) technology of moon probe is becoming more important than ever. Because of the communication delay induced by the large distances between the earth and moon, human kinds are hardly able to guide probe to landing safely on the moon. Probe will have to use on-board sensors and algorithms. However, current navigation technology can't provide precision motion estimation for probe landing control system[2]. Therefore, new motion estimation technology must be developed. Computer vision may offer an approach to solve this problem.

Up to now, computer vision has been successfully used for autonomous navigation of robot and Micro Air Vehicle [3], autonomous landing of Unmanned Helicopter[4], and intelligence traffic[5].

Some appliances are being researched in the field of space exploration. The Near Earth Asteroid Rendezvous (NEAR), rendezvoused with asteroid Eros 433 in February 2000, used optical navigation extensively for orbit determination and small body 3-D modeling[6]. The Deep Space 1 mission as a part of the New Millennium Program is flying an autonomous optical navigation technology demonstration. The DS-1 AutoOpNav system used onboard centroiding of reference asteroids for autonomous navigation during small body fly-bys [7]. However, for probe accurate landing, the new, more robust and more precise vision algorithm should be developed.

Image processing is the first step and a essential part of computer vision navigation. Generally, image processing of vision navigation study is mainly focused on feature detection and tracking. In 1997, Kawaguchi, from Japan, created a method, in which several fixed reference windows and a 2-D fast Fourier transform (FFT) algorithm were used in the correlation tracking[8]. In many case, flat areas must be chosen for landing security. Safe Landing Area Determination (SLAD),which offered by A. Johnson from Jet Propulsion Laboratory(JPL), achieved this task successfully [9]. However, there are no features or distinctive features on those landing sites. Although tracking by feature points is not the best way, feature points haved to be filtered in order to avoid this problem. Point tracking is the usual way in vision navigation .In 1999, Andrew E. Johnson, from JPL, generated a method based on automatic feature point(FP) tracking between a pair of images followed by two frames[2]. As a part of MUSES-C, Japanese scientist Toshihiko Misu offered an algorithm based on fixation area, which fixates on the areas with local varying intensity, and employs shading pattern as a fixation area, and then tracked the center of the fixation area[10]. Real-time image processing is the essential request in computer vision navigation. Toshihiko's method requires less computation than Johnson's. Nevertheless, the Number of fixation area's center is difficult to satisfy the requirement of following navigation algorithm. This paper introduces an image processing algorithm using feature point tracking in probe autonomous landing which built on the methods from Johnson and Toshihiko. The procedure is as follow (Fig1):

Step1. Extract fixation areas as templates and sub-images.

Step2. Match templates and detecting FPs from sub-images

Step3. Match FPs

In section 2, fixation area detection algorithm is proposed. Feature point detection and tracking in the sub-images are explained in section 3 and section 4. Computer simulations of this algorithm are illustrated to demonstrate this image processing algorithm in section 5. In Section 6 we state conclusions.

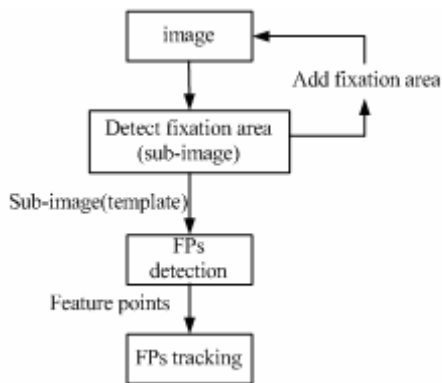

**Fig.1 Procedure of algorithm**

## II. FIXATION AREA DETECTION

In artificial environment, edges and vertexes are usually used as features in image tracking. But in natural environment, these shapes can't be found easily. Paper[11] proposed the use of ground control points based on Gabor function in image registration of natural terrain. In paper[10], shading pattern is employed as a fixation-area. To reduce computation, the averaging, Laplacian filtering and variance are used. In this paper, improved Toshihiko Misu's method is used firstly to pre-process primary image for obtaining sub-image. The purpose of this step is that block matching is more robust than point matching. We can match block firstly in entire image and then match points in sub-image.

The template for matching block is extracted as follows[10]:

Step1. Enhance specific spatial wavelength of the original image by 2-D band-pass-filter.

Step2. Calculate local variances of filtered image to evaluate "contrast".

Step3. Extract high local-variance areas as templates.

Step4. Least-square block matching. The matching is performed so as to minimize square error of the intensity between templates and region to be matched.

**2.1 Band-Pass-Filter**

The conception of filtering comes from the Fourier transform for signal processing in the frequency domain. Image processing is interested in filtering operations that are performed on the pixels of a digital image. The filter in the image processing is called spatial filtering. In generally, the image filter can be divided into smoothing spatial filter and sharpening spatial filter.

Firstly, original image should be processed by smoothing spatial filter to reduce noise. In fact, smoothing spatial filter is a type of low-pass-filter(LPF), includes linear filter, such as averaging filters, and nonlinear filter, such as median filters. To reduce the consumption of computational power and storage, we use averaging as LPF, which is different from averaging filter. The averaging filter can't change the size of original image and spend more time. The equation of averaging is:

$$E_s(u,v) = \frac{1}{S_x S_y} \sum_{i=0}^{S_x-1} \sum_{j=0}^{S_y-1} E(S_x u + i, S_y v + j) \quad (2.1)$$

Where $E_s(u,v)$ is the intensity of averaged and sub-sample image at $(u,v)$. The size of this image is $\frac{1}{S_x S_y}$ of size of original image. ($S_x, S_y$) is the sub-sampling interval.

Secondly, the 8-neighbor Laplacian is used as high-pass-filter. The mask of Laplacian as follows:

| 1 | 1 | 1 |
|---|---|---|
| 1 | -8 | 1 |
| 1 | 1 | 1 |

**2.2 Variance Map**

To show locally varying intensity, the variance map is calculated. Statistical variance within a window and the size of which is equal to computed template is computed. The formulation is:

$$V(u,v) = \frac{S_x S_y}{W_x W_y} \sum_{i=0}^{w_x-1} \sum_{j=0}^{w_y-1} \{E_L(u+i,v+j)\}^2 - \{\frac{S_x S_y}{W_x W_y} \sum_{i=0}^{w_x-1} \sum_{j=0}^{w_y-1} E_L(u+i,v+j)\}^2$$

Where $V(u,v)$ is the local variance of Laplanian-filtered image. $W_x \times W_y$ is the size of template. $E_L(u,v)$ is the intensity of Laplanian-filtered image at $(u,v)$. $w_x = \frac{W_x}{S_x}, w_y = \frac{W_y}{S_y}$.

**2.3 Template extraction**

Templates for tracking are extracted according to variance map. The high-local-variance points are selected as the template. We should pay attention: a. The points chosen as template in the variance map are not near the edge of the map, in order to easily track points in next frame image. b. Some other templates which aren't near extracted ones should be extracted also. This is because the cluster or single template would be weak to the observation noise which includes tracking error and error from the range measurement.

**2.4 Template matching**

To track the fixation area, we employ least-square block matching. The matching is performed so as to minimize the square error of the intensity between template and region to be matched. The number of templates should be certain. If any template which has been out of the image can't be tracked, the new template is extracted instead.

### III. FP DETECTION FROM FIXATION AREA

To best our knowledge, some real-time FPD algorithms have been described in many literatures. FPD based on Harris Corner Finding algorithm has been employed by ASSET-2 motion, segmentation and shape tracking system[12]. After detecting the fixation area, paper[10] assumed the center of the fixation area as an feature point. I don't think this is a good idea. In this paper, a FPD algorithm is selected to detect feature points from fixation area. We can use those points for match.

In this paper, we select and improve a FPD algorithm of Benedetti and Perona[14], which is based on the work of Tomasi and Kanada[13]. There are two reasons to select this algorithm.

a) This algorithm decreased the complexity of Tomasi's method by eliminating the demand for any transcendental arithmetic operations and didn't require calculation of square root.

b) This algorithm is easy to track across several frames.

It is difficult to detect and an individual point because of the presence of noise and distortion. For this reason patches of pixels are actually considered. The method proposed in paper[14] for detect FP is expressed simply as follows:

Step1. Computing the image gradient across the image at each pixel location

$$I_x(x,y,t) = \frac{I(x+1,y) - I(x-1,y)}{2} \quad (3.1)$$

$$I_y(x,y,t) = \frac{I(x,y+1) - I(x,y-1)}{2} \quad (3.2)$$

$I(x,y,t)$ denoted the continuous function defining the brightness values of a sequence of images captured by a camera. The partial derivatives of $I(x,y,t)$ with respect to x, y, are denoted respectively by $I_x(x,y,t)$ and $I_y(x,y,t)$.

Step2. Defining matrix G

$$G = \begin{bmatrix} \sum_{W_1}(I_x(x,y,t))^2 & \sum_{W_1} I_x(x,y,t)I_y(x,y,t) \\ \sum_{W_1} I_x(x,y,t)I_y(x,y,t) & \sum_{W_1}(I_y(x,y,t))^2 \end{bmatrix}$$

$$= \begin{bmatrix} a & b \\ b & c \end{bmatrix} \quad (3.3)$$

W1 denoted feature patch of pixels including pixel point(x,y).

Step3. Find $P_{\lambda_t(i,j)}$

$$P_{\lambda_t(i,j)} = (a - \lambda_t)(c - \lambda_t) - b^2 \quad (3.4)$$

Step4. retain pixel if $P_{\lambda_t(i,j)} > 0, a(i,j) > \lambda_t$, $\lambda_t$ is a threshold value.

Step5. Perform a final non-maximum suppression step on $\lambda_1(i,j)$, yielding the actual feature locations.

The fixation is the area with local varying intensity, so good features for tracking are expected to be abundant. At present, there are two strategies for feature point search: exhaustive search and random search strategy. The first one is traditional feature point detection algorithm which searches the image for every distinction exhaustively. Suppose that N feature points are needed for motion estimation which is the purpose of searching. The random search strategy selects a pixel at randomly from the image. If the randomly selected pixel has an interest value greater than a predetermined threshold, it is as a feature point. This procedure is repeated until N feature points are detected. Compared with the first algorithm, the second one raised the speed of detection. In this paper, we use exhaustive search to obtain enough FPs because the size of sub-image is small.

### IV. FEATURE POINT TRACKING

The next step is to locate the features detected in the first frame in the second frame. This procedure is called feature tracking. Feature tracking can be divided into two groups of algorithm: correlation based methods and optical flow based method. Correlation based methods require more computation and are appropriate when the motion of features in the image is large, Such as RANSAC. In this paper, we use Tomasi-Kanade's feature tracking which is optical flow based method[15].

When the camera moves, the patterns of image intensities change in a complex way. However, images taken at late time instants are usually strongly related to each other, because they refer to the same scene taken from only slightly different viewpoints.

$$I(x,y,t) = I(x+d_x, y+d_y, t+\tau) \quad (4.1)$$

Using Taylor-series expansion, obtaining:

$$I(x+d_x, y+d_y, t+\tau) \approx I(x,y,t) + I_x d_x + I_y d_y + I_t \quad (4.2)$$

From (4.1) and (4.2):

$$I_x d_x + I_y d_y + I_t = 0 \quad (4.3)$$

Equation(4.3) is called optical flow constraint equation.

If every pixel(N pixels) in the feature window has the same displacement, we can obtain:

$$\begin{bmatrix} I_x^1 & I_y^1 \\ \bullet & \bullet \\ \bullet & \bullet \\ \bullet & \bullet \\ I_x^N & I_y^N \end{bmatrix} \begin{bmatrix} d_x \\ d_y \end{bmatrix} = - \begin{bmatrix} I_t^1 \\ \bullet \\ \bullet \\ \bullet \\ I_t^N \end{bmatrix} \quad (4.4)$$

The displacement of window can be gained by least-square.

$$\begin{bmatrix} \sum_{k=1}^{N}(I_x^k)^2 & \sum_{k=1}^{N} I_x^k I_y^k \\ \sum_{k=1}^{N} I_x^k I_y^k & \sum_{k=1}^{N}(I_y^k)^2 \end{bmatrix} \begin{bmatrix} d_x \\ d_y \end{bmatrix} = - \begin{bmatrix} \sum_{k=1}^{N} I_x^k I_t^k \\ \sum_{k=1}^{N} I_y^k I_t^k \end{bmatrix} \quad (4.5)$$

i.e., $Gd = e \quad (4.6)$

Displacement is: $d = G^{-1}e \quad (4.7)$

### V. COMPUTER SIMULATIONS AND RESULT

According to the algorithm which has been expounded in this paper, we give the result of computer simulations in this section. Fig 2(a)(b)(c) are original images of continuous frame.

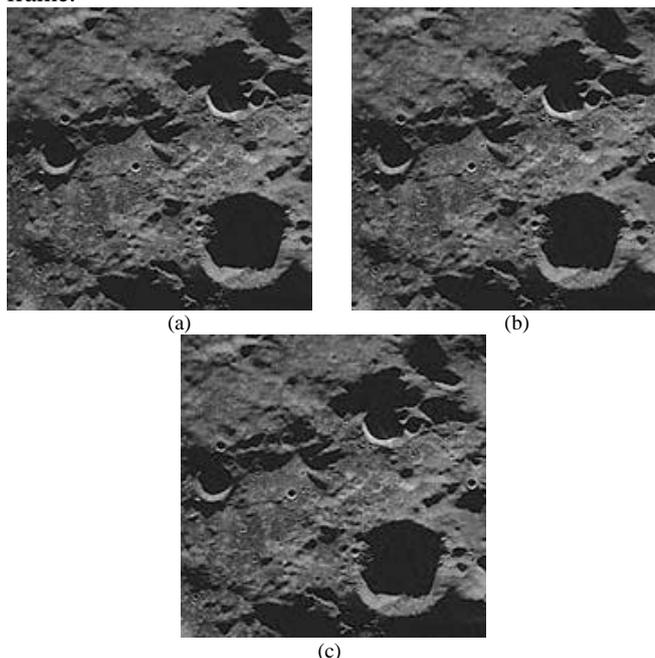

**Fig2. original images of continuous frame**

Fig2(a) is LPF image of Fig1(a), $S_x = S_y = 5$. Fig2(b) is the image by Laplacian filtering. Fig3(c) is variance map(4×4) of Fig1(a). Fig3(d) is the extracted areas on original image.

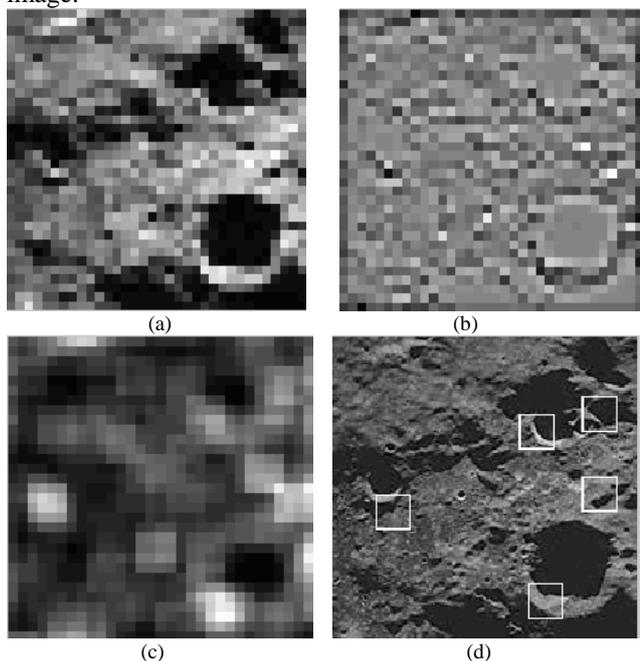

**Fig3.proceed image**

Fig4(a)(b)(c)(d)(e) is five templates(20×20) from Fig2(a)

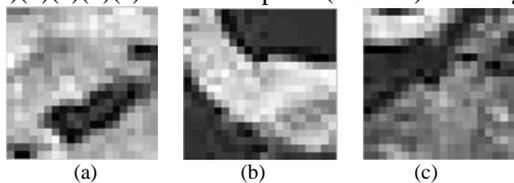

**Fig4.templates(sub-image)**

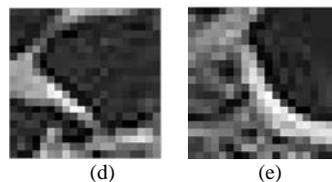

**Fig4.templates(sub-image)**

Fig5 illustrates the result of template match. White Square means the position of sub-images from 3 frames images in the first frame. Arrow means the direction of template moving.

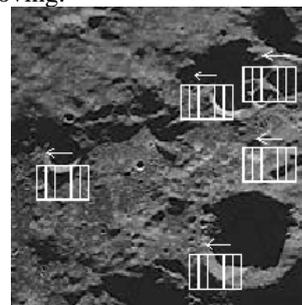

**Fig5.template match**

Fig6 illustrates the results of FPs detection. W1=3, threshold=1500. Fig7 illustrates the result of FPs tracking in the sub-image. Fig8 is the result of FPs tracking in the whole image.

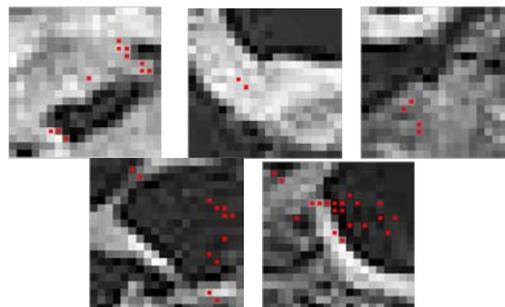

**Fig6. FPs detection**

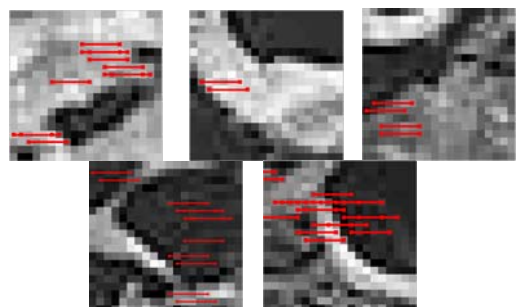

**Fig7. FPs tracking of sub-image**

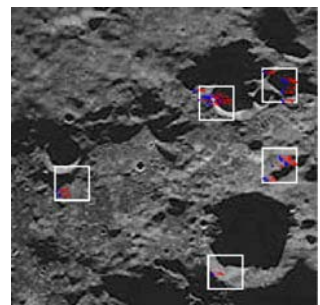

**Fig8. FPs tracking of whole image**

## VI. Conclusions

Computer simulation demonstrates that we can receive enough FPs real-timely and correctly. FPs can be extracted from the sub-images, as a result, FPs tracking algorithm cost less computation than other methods. At the same time, the field of FPs tracking becomes smaller, so the level of FPs matching becomes higher. This algorithm should be improved in the future. We should extend the distance between different feature points. The purpose is to make the result of motion estimation more correct